\documentclass[10pt,twocolumn,letterpaper]{article}

\PassOptionsToPackage{table,xcdraw}{xcolor}

\usepackage[pagenumbers]{iccv}

\usepackage{graphicx}
\usepackage{amsmath}
\usepackage{amssymb}
\usepackage{booktabs}
\usepackage{bbm}
\usepackage{multirow}
\usepackage{comment}
\usepackage[normalem]{ulem}
\usepackage{nicefrac}
\usepackage{mathtools}
\usepackage{algorithm}
\usepackage{algpseudocode}
\usepackage{pifont}
\usepackage{makecell}

\newcommand{\encoder}{E_\xi}

\newcommand{\decoder}{D_\gamma}
\newcommand{\decoparams}{\gamma}
\newcommand{\model}{\epsilon_\theta}
\newcommand{\graph}{\mathcal{G}}
\newcommand{\hatgraph}{\hat{\mathcal{G}}}

\newcommand{\sequence}{\mathbf{Z}}

\newcommand{\dimnode}{d_\text{node}}
\newcommand{\dimedge}{d_\text{edge}}
\newcommand{\dimlatent}{C}
\newcommand{\winsize}{S}

\newcommand{\jsim}{J_\text{sim}}
\newcommand{\jdist}{J_\text{dist}}
\newcommand{\jsimb}{\mathbf{J}_{\mathbf{\text{sim}}}}

\newcommand{\expec}{\mathbb{E}}

\newcommand{\cmark}{\ding{51}}
\newcommand{\xmark}{\ding{55}}

\definecolor{comments}{rgb}{0.13, 0.55, 0.13}
\definecolor{keywords}{rgb}{0,0,0.75}
\definecolor{strings}{rgb}{0.75,0,0}
\usepackage{lipsum}
\definecolor{darkpurple}{RGB}{75, 0, 130}

\definecolor{iccvblue}{rgb}{0.21,0.49,0.74}
\usepackage[pagebackref,breaklinks,colorlinks,allcolors=iccvblue]{hyperref}

\def\paperID{XXXXX}  
\def\confName{ICCV}
\def\confYear{2025}

\newcommand{\ours}{FORESCENE\xspace}
\title{FORESCENE: FOREcasting human activity via latent SCENE graphs diffusion}

\author{Antonio Alliegro$^{1}$
\quad
Francesca Pistilli$^{1}$
\quad
Tatiana Tommasi$^{1}$
\quad
Giuseppe Averta$^{1}$
\and
$^{1}$ Politecnico di Torino\\
{\tt\small firstname.lastname@polito.it}\\
}

\begin{document}
\maketitle
\begin{abstract}
Forecasting human-environment interactions in daily activities is challenging due to the high variability of human behavior. 
While predicting directly from videos is possible, it is limited by confounding factors like irrelevant objects or background noise that do not contribute to the interaction. 
A promising alternative is using Scene Graphs (SGs) to track only the relevant elements. However, current methods for forecasting future SGs face significant challenges and often rely on unrealistic assumptions, such as fixed objects over time, limiting their applicability to long-term activities where interacted objects may appear or disappear. 

In this paper, we introduce \ours, a novel framework for Scene Graph Anticipation (SGA) that predicts both object and relationship evolution over time. 
\ours encodes observed video segments into a latent representation using a tailored Graph Auto-Encoder and forecasts future SGs using a Latent Diffusion Model (LDM). 
Our approach enables continuous prediction of interaction dynamics without making assumptions on the graph’s content or structure. 
We evaluate \ours on the Action Genome dataset, where it outperforms existing SGA methods while solving a significantly more complex task.
\end{abstract}

\section{Introduction }
\label{sec:intro}

\begin{figure}
    \centering
    \includegraphics[width=1.0\linewidth]{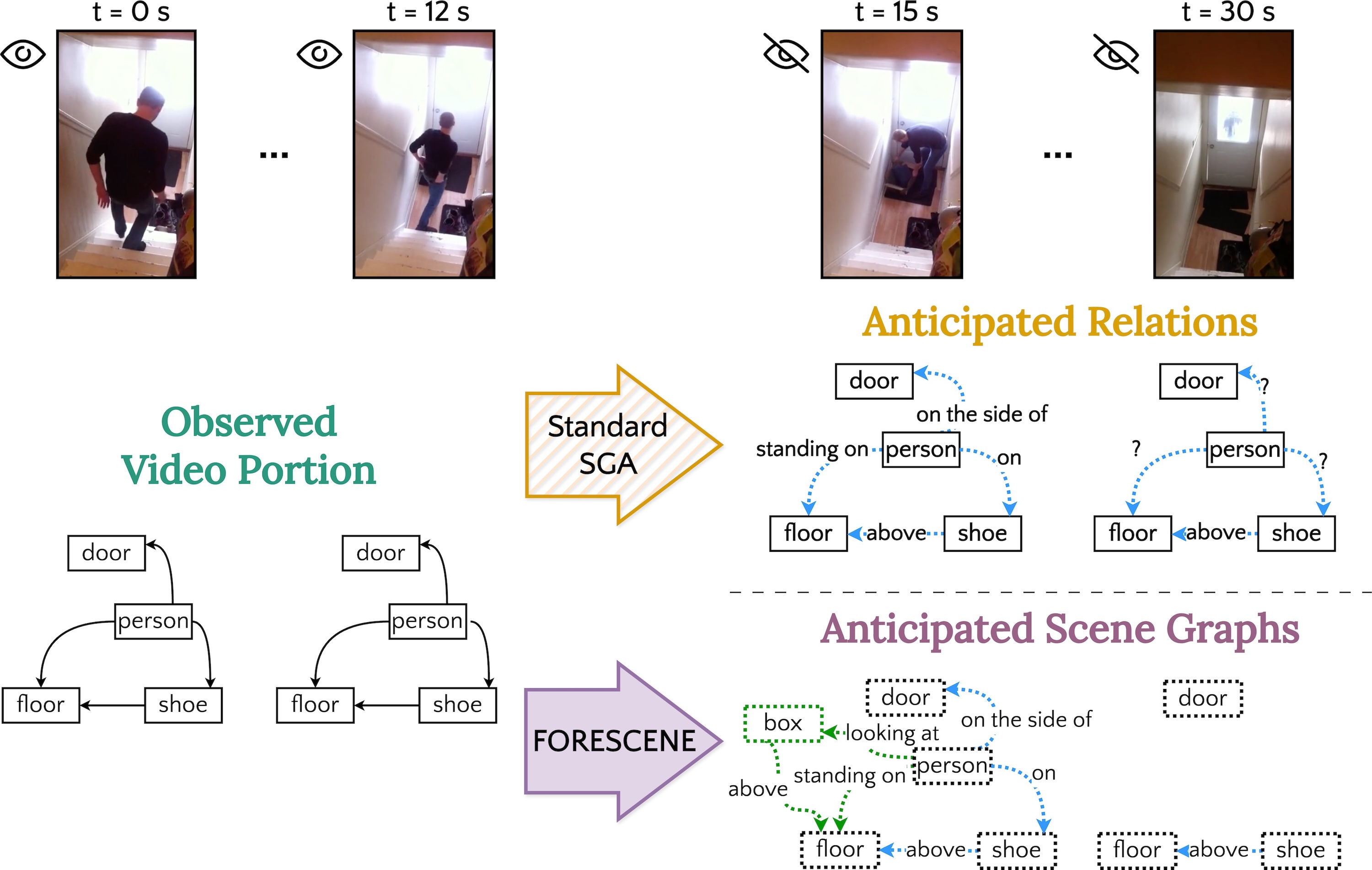}
    \vspace{-6mm}
    \caption{
    Scene graphs for human-environment interactions represent actors and objects as nodes, and relationships as edges. Dotted lines indicate predictions, while solid lines denote fixed elements. A model forecasting these interactions should predict the entire graph -- both nodes and edges -- allowing objects to appear and disappear over time. Unlike other methods, which only update relationships while keeping object nodes fixed, \ours achieves complete (nodes+edges) graph forecasting.\label{fig:teaser}}
    \vspace{-4mm}
\end{figure}

Imagine a human and a service robot sharing the same workspace. To favour a safe and effective collaboration, the robot should be able to reason about the current human activity, and potentially forecast its evolution, including the interaction of the human with the objects around. For example, let's assume the human is preparing a cup of coffee. The robot should build a clear representation of the actor and the objects participating in such interaction, such as the cup, the coffee machine and a bottle of water. 
Using \textit{images} to feature such information is in principle feasible, yet it comes with many confounding factors irrelevant to the task \cite{lai2025lego}, and raises the question of how to encode relations in visual form. 
Conversely, \textit{text} offers a compact representation of relevant information but can be ambiguous, leading to interpretation challenges, without considering that visual grounding of textual descriptions is a challenge per-se \cite{lai2025lego}. 
A trade-off encompassing the advantages of both would be through the adoption of \textit{Scene Graphs} (SGs) \cite{ji2020action}, which represent the actor and objects as nodes and their relationships as edges. 

After making coffee, the human takes his cup and reaches the desk to write a few notes on a piece of paper. The representation made by the robot should keep updating the relation between the human and the cup, drop the one with the coffee machine, and instantiate a few new nodes featuring the paper, the chair, and the desk. 
A robot that effectively reasons about human behavior and plans its actions accordingly, should not only be able to predict a scene graph from the observed video frames (Scene Graph Generation - SGG) but also forecast how it will evolve over time. We refer to this task as Scene Graph Anticipation (SGA).

Scene graphs (SGs) are commonly used to encode knowledge from various domains (e.g., images, point clouds, protein structures) into a compact representation for many downstream tasks \cite{Li2024surveySGG,Chang2023surveyPAMI}. 
However, forecasting SGs poses unique challenges. Unlike images, which follow a regular grid, graphs are sparse and unstructured, complicating learning. 
This complexity further increases when SGs are used to describe human-environment dynamic interactions, where involved objects may appear and disappear over time, causing sudden variations in the number of nodes and edges.

Predicting the full temporal evolution of such graphs is not trivial, and prior work on SGA \cite{peddi2024towards} simplifies the task by focusing solely on forecasting edge attributes (\ie, relationships) while keeping the set of nodes (\ie, objects) fixed. However, this constraint severely limits the applicability of these approaches, making them unsuitable for modeling long-term activities (see Fig.~\ref{fig:teaser}).

In this work, we extend SGA to forecast complete scene graphs by introducing \ours that operates in two stages: i) a tailored Graph Auto-Encoder maps the observed video portion into a latent representation capturing co-occurrence of objects and their task-related functionality; ii) a Latent Diffusion Model predicts the temporal evolution of scene graphs conditioned on the seen portion. 
This approach enables the anticipation of the objects involved in the activity and their pairwise relationships without imposing constraints on the graph’s content or structure.

To fully expose the challenges of real-world use cases, we also define a new experimental benchmark built on the Action Genome dataset \cite{ji2020action} with settings of increasing difficulty in terms of objects distribution shift between the observed and future video portions. 
We assess the performance of \ours and several baselines on two primary tasks: i) Object discovery, measuring accuracy in forecasting the involved objects at each activity step; and ii) Relation discovery, evaluating the prediction of relationships between objects involved in the interaction. 
Our thorough experimental analysis shows that \ours excels at relation discovery and overcomes the competitors in forecasting the appearance and disappearance of objects in unconstrained settings.

In a nutshell, our technical contribution to solve the SGA task can be summarized as follows: 
\begin{itemize}
    \item Graph Auto-Encoder~(GAE): We propose a novel encoding-decoding framework that maps unstructured graphs to a smooth, continuous latent representation and reconstructs explicit scene graphs (objects + relations) without constraints on node and edge counts. 
    \item Latent Diffusion Model~(LDM): 
    We complement the Graph Auto-Encoder with a Latent Diffusion Model specifically tailored for scene graph anticipation.
    \item By combining our GAE and LDM, we introduce \ours, the first framework for a comprehensive and temporally consistent prediction of the dynamics of human-environment interaction as sequences of scene graphs, without making assumptions on the content or structure of the predicted graphs.
    \item We evaluate \ours’s ability to jointly forecast objects and their relationships on the Action Genome dataset \cite{ji2020action}, achieving top results on the standard SGA~\cite{peddi2024towards} benchmark and on our proposed SGA testbed under object distribution shift.
\end{itemize}

\section{Related works}
\label{sec:related}

\subsection{Scene Graph Generation and Anticipation}
Extracting graphs that encode key elements of a scene - typically named \textbf{Scene Graph Generation (SGG)} - has a longstanding history \cite{Li2024surveySGG,Chang2023surveyPAMI} and has proven to be a relevant tool for image retrieval \cite{JJohnson2015retrieval}, visual question answering \cite{Teney2017graphVQA}, image generation \cite{scenegraphgen2021}, as well as for robot planning and navigation \cite{Amodeo2022Access,Amiri2022RAL,Kim2020Cybernetics}. 
Recent studies have explored the use of scene graphs to represent human activities \cite{ji2020action, rai2021home}. The seminal work of Action Genome \cite{ji2020action} catalyzed a growing body of research focused on developing SGs for human-environment interactions. In this context, nodes encode objects' geometric and semantic properties, while edges represent functional, spatial, or contact-based relationships (\ie, predicates) between them. 
Stemming from Visual Genome \cite{krishna2017visual}, several works contributed to SGG in 2D \cite{cong2023reltr} and 3D \cite{kim20193}, also considering the potential advantage of including foundation models for open-vocabulary \cite{chen2023expanding}, panoptic \cite{zhou2023vlprompt}, zero/few-shot or weak supervision \cite{li2024zero,kim2024llm4sgg,zhao2023less}.

For human activity understanding, the temporal dimension (\ie, the dynamics) of the action plays a crucial role. Numerous studies extended SGG to incorporate dynamic aspects, resulting in \textbf{Spatio-Temporal Graphs}. In this case, the graph still features objects in nodes, and their relation through edges, but their features may vary in time to account for the temporal dimension in video-level SGG \cite{pu2023video, teng2021target,yang2023panoptic, cong2021spatial, yang20244d, nguyen2024hig}. Videos also come with several additional challenges related to dataset unbalance of relations \cite{nag2023unbiased,wu2023weakly} or noise and blur in certain frames \cite{lin2024td2}.

In many practical scenarios, it is required to reason about activities without full access to the complete video, for instance because the action is still ongoing. This requires models able to perform \textbf{Scene Graph Anticipation (SGA)}, \ie to predict future instances of a temporal graph conditioned to a set of frames available (observed portion). While several works have attempted to forecast future actions \cite{wang2023pdpp} and future scene realizations (\ie future frames) conditioned to activities \cite{lai2025lego} --- mostly leveraging the extensive literature in video generation --- little has been done to predict future instances of SGs. The authors of \cite{peddi2024towards} introduced the SGA task, discussing the relevance of predicting the temporal evolution of scene graphs for downstream tasks such as action anticipation or anomaly detection. 
The same work proposed a strategy to learn the temporal evolution of relations between objects (\ie, predicates on edges) but assume \textit{objects continuity} for the whole rollout (\ie, objects in the graph are fixed to those observed in the last seen frame). 
However, such an assumption is strongly limiting for many practical scenarios, where objects or tools may appear or disappear during the video rollout or even change state (like the flour becoming dough while making pizza). 

In this regard, \ours reframes the SGA task, fully addressing the dynamic nature of human activities by predicting both the evolution of predicates and the appearance or disappearance of objects.

\subsection{Latent Generative models}
Diffusion Models (DM) demonstrated impressive abilities in text-conditioned image and video synthesis \cite{rombach2022high, blattmann2023stable, meng2021sdedit}. 
They rely on an iterative process that perturbs data by adding noise, and a deep neural network trained to progressively revert this process. 
At inference time, the model begins with arbitrary input noise and progressively refines it through iterative denoising, ultimately generating a meaningful sample that aligns with the training data distribution while remaining distinct from it.
Interestingly, DMs demonstrated notable capabilities in fitting human motion \cite{yuan2023physdiff}, even in the case of complex activities \cite{wang2023pdpp}. 

Recent works improved generative model efficiency by leveraging latent-space generation strategies~\cite{rombach2022high,podell2024sdxl}. A Variational Auto-Encoder (VAE)~\cite{Kingma2014} first maps data samples to a latent space, where a diffusion process models them. The generated latents are then decoded back to the original data space. This approach is particularly suited for high-resolution images and videos, and its performance hinges on the VAE’s capacity to encode data into a rich latent space and reconstruct generated latents with high fidelity. 
It was also adopted to generate molecules represented as 3D point clouds \cite{xu2023geometric}: the basic assumption is that the number of points and their nature is known and the model only learns to change their spatial positioning. 

With \ours, we tackle SGA by introducing a specialized latent graph generative model. To the best of our knowledge, this is the first work to use latent diffusion for forecasting human activity dynamics as a sequence of scene graphs, without imposing restrictive conditions on the structure of the forecasted graphs. 

\section{Method}
\label{sec:method}
\begin{figure*}[t]
    \centering
    \includegraphics[width=1.0\linewidth]{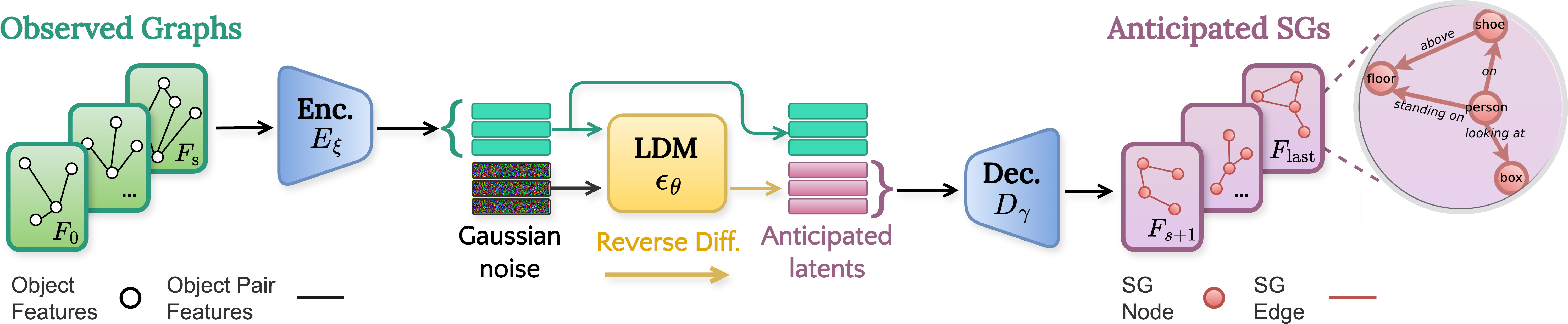}
    \caption{Overview of the proposed method at inference time to solve the task of Scene Graph Anticipation. 
    Observed frames $\{F_0 ...F_s\}$ are encoded into latent representations using the Graph Encoder. 
    Latent codes are then fed into the diffusion model as conditioning input to anticipate the future unseen latents. 
    The anticipated latents are finally transformed back by the decoder into a sequence of complete (objs + rels) scene graphs $\{F_{s+1} ... F_\text{last}\}$.\label{fig:method}}
    \vspace{-4mm}
\end{figure*}

Our goal is to design a model able to predict future instances of scene graphs from a partial video observation. 
The videos depict activities of humans interacting with multiple
objects simultaneously (\eg sitting on a chair while writing on paper with a pen). 
We tackle this problem by first encoding scene graphs extracted from observed RGB frames into a fixed-size latent representation through a tailored Graph Auto-Encoder (GAE), and then using a Latent Diffusion Model (LDM) to generate future instances of scene graphs conditioned on the observed. 
By training the LDM on sequences of graph latents, we capture the temporal evolution of scene graphs and patterns of human activity in everyday scenarios. 
At test time, the LDM generates temporally coherent latent codes representing future graphs, which are then decoded back into complete scene graphs by the GAE’s decoder (see Fig.~\ref{fig:method}). 
This process results in a sequence of scene graphs that model the temporal evolution of the human-environment interaction. 
An inherent advantage of using our LDM is its ability to generate multiple roll-outs by varying the diffusion seed, enabling diverse plausible forecasts of human-environment interactions. 
\ours leverages this capability to propose multiple alternative solutions, capturing the variability of human behavior in daily activities.

The following sections provide an in-depth description of each step and a detailed formulation of our approach.

\subsection{Notation and Task Definition}
We define a scene graph as $\mathcal{G} = \{\mathcal{V}, \mathcal{E}\}$, where $\mathcal{V}$ represents the set of nodes, each corresponding to an object, and $\mathcal{E}$ denotes the set of edges, capturing the relationships between nodes. 
Each node $v_i \in \mathcal{V}$ is associated with an object category label $v_i^c \in \mathcal{C}$ and bounding box coordinates $v_i^b \in [0,1]^4$. 
An edge $e_{ij} \in \mathcal{E}$ serves to model the relationship between nodes $v_i$ and $v_j$, with $v_i$ acting as the \textit{subject} and $v_j$ as the \textit{object} of the relation. The relationship category is represented by $p^c_{ij} \in \mathcal{P}$. 

\textbf{Scene Graph Generation (SGG)} predicts a scene graph $\mathcal{G}$ from a static image or a video segment $I$. This task requires solving both object detection and relation extraction problems. Instead, in \textbf{Scene Graph Anticipation (SGA)} the model predicts scene graphs $\mathcal{G}_{F_{s+1}}, \ldots, \mathcal{G}_{F_{last}}$ for future unseen frames $\{I\}_{F_{s+1}}^{F_{last}}$, based on a partial video observation $\{I\}_{0}^{F_s}$. 
In the SGA formulation, we ask the model to anticipate the emergence or disappearance of new objects (\ie, nodes), as well as to figure out their relationship (\ie, edges), thereby predicting future graph structures that align with the temporal dynamics of human-environment interactions in the video. 

\subsection{Graph Auto-Encoder}
\label{subsec:stageone}
The first stage of our method is a GAE that maps observed frames into fixed-size latent graph representations and decodes them into scene graphs. 
We utilize a Graph Convolutional Network (GCN) encoder and a transformer-based decoder~\cite{vaswani2017attention}. 
Both modules are specifically designed to handle the inherent variability of scene graphs modeling human-environment interactions, such as varying numbers of nodes and edges.

\noindent\textbf{Graph Encoder $(\mathcal{G} \overset{\encoder}{\longrightarrow} \mathbf{z})$} 
The encoder learns to map a graph $\mathcal{G} = \{\mathcal{V}, \mathcal{E}\}$ to a latent code $\mathbf{z}$. Input graphs are populated with node features $\{\phi_{v,i}\}_{i=1}^{|\mathcal{V}|}$ and edge features $\{\phi_{e,ij}\}_{e=1}^{|\mathcal{E}|}$. 
Following \cite{cong2021spatial, feng2023exploiting, peddi2024towards}, \textit{node features} $\phi_{v,i}$ are initialized using object features $v_i^f$ extracted from a frozen object detector~\cite{faster_rcnn} and bounding box coordinates $v_i^b$. 

Specifically, we obtain $\phi_{v,i}$ as: 
\begin{equation}
    \phi_{v,i} = \text{Concat}(W_1 v^f_{i}, W_2 v^b_i),
\end{equation}
where $W_1$, $W_2$ are learnable weight matrices. 

For \textit{edge features} $\phi_{e,ij}$, we follow \cite{cong2021spatial}, and concatenate visual and semantic features of object pairs: 
\begin{equation}
    \phi_{e,ij} = \text{Concat}(W_3 \phi_{v,i}, W_4 \phi_{v,j}, W_5 U_{ij}, \mathbf{v}_i^c, \mathbf{v}_j^c),
\end{equation}
where $W_3$, $W_4$, and $W_5$ are learnable weight matrices, $U_{ij}$ represents the processed feature maps of the union box computed via RoIAlign~\cite{faster_rcnn}, and $\mathbf{v}_i^c$ and $\mathbf{v}_j^c$ are learnable semantic embedding vectors for the object categories. 
Note that the graph $\mathcal{G}$ input to the encoder does not include relationship categories and can be constructed directly from object annotations or with an off-the-shelf object detector.

The node and edge features are then structured as triplets $t_{ij} = \langle \phi_{v,i}, \phi_{e,ij}, \phi_{v,j} \rangle$ and fed into the GCN architecture proposed in \cite{johnson2018image}. The GCN iteratively refines the triplet features $t_{ij}$ and propagates information across the graph:
\begin{equation}
    \phi_{v,i}^{l}, \phi_{e,ij}^{l}, \phi_{v,j}^{l} = \text{G}(\phi_{v,i}, \phi_{e,ij}, \phi_{v,j}),
\end{equation}
where $\text{G}(\cdot)$ represents the GCN, and the superscript $^{l}$ denotes the refined features at the $l$-{th} layer. 
The output of the last GCN layer consists of node features $\phi_{v} \in \mathbb{R}^{\dimnode}$ (with dim. $|\mathcal{V}| \times \dimnode$) and edge features $\phi_{e} \in \mathbb{R}^{\dimedge}$ (with dim. $|\mathcal{E}| \times \dimedge$). 
These features are then max-pooled (across the node and edge dimensions), concatenated, and linearly projected to obtain the latent graph representation $\mathbf{z} \in \mathbb{R}^{\dimlatent}$.

To encourage the encoder to learn semantic features for nodes and edges and store them in the latent $\mathbf{z}$, before max-pooling, we fed $\phi_v$ and $\phi_e$ into two auxiliary classification heads to predict node and edge categories.

We then use the following loss function:
\begin{equation}
    \label{eq:enco_loss}
    \mathcal{L}_\text{enco} = \mathcal{L}_\text{nodes} + \mathcal{L}_\text{edges}.
\end{equation}
Here, $\mathcal{L}_\text{nodes}$ and $\mathcal{L}_\text{edges}$ are cross-entropy losses for node and edge classification, respectively (details in supplementary).  

\noindent\textbf{Graph Decoder $(\mathbf{z} \overset{\decoder}{\longrightarrow} \hat{\mathcal{G}})$}
Inspired by \cite{carion2020detr}, our decoder $\decoder$ processes the graph latent representation $\mathbf{z}$ alongside $N$ object queries to generate a feature representation of each node in the scene graph. 
The architecture is composed of $L$ stacked blocks, each of which sequentially performs cross-attention, self-attention, and feed-forward operations. 
In the $l^{th}$ block, the interaction of object queries $h^{l}$ with $\mathbf{z}$ through cross-attention produces $h^{l}_{\mathbf{z}}$. 
This interaction is followed by multi-head self-attention, which computes the attention matrix
\begin{equation}
    A_{h}^{l}= \text{softmax}\left(Q_{h}^{l} {K_{h}^{l}}^T \slash {\sqrt{d_{\text{head}}}}\right),
\end{equation}
where $A_{h}^{l} \in \mathbb{R}^{N \times N}$ represents the attention weights, and $Q_{h}^{l}, K_{h}^{l} \in \mathbb{R}^{N \times d_{\text{head}}}$ denote the respective queries and keys. 
Each block then processes the attended object queries through a feed-forward network, producing the input $h^{l+1}$ for the subsequent block. 

The output of the last decoder block $h^{L}$ is finally fed to $\text{MLP}_\text{objects}$ and $\text{MLP}_\text{boxes}$, which respectively output the predicted (indicated with \emph{hat} notation) object category distribution over the $\mathcal{C}$ object classes, $\hat{v}^c \in \mathbb{R}^{N \times \mathcal{C}}$,  and bounding boxes $\hat{v}^b \in \mathbb{R}^{N \times 4}$ for each object query. 

For \textit{edge prediction}, we follow \cite{im2024egtr}. The queries $Q_h^{l}$ and keys $K_h^{l}$ are interpreted as subjects and objects of the relationships, respectively. 
These are concatenated across heads to form $Q^{l}$ and $K^{l}$, processed by layer-specific MLPs to generate relational vectors $R^{l}$:
\begin{equation}
    R^{l} = \left[Q^{l}W^{l}_S; K^{l}W^{l}_O\right],
\end{equation}
where $R^{l} \in \mathbb{R}^{N \times N \times 2d_{\text{model}}}$ and $W^{l}_S, W^{l}_O$ are learnable weights specific to each block. 
Then, we adopt the same gating mechanism from \cite{im2024egtr} to obtain the refined predicted relational features $\hat{\phi_{e}} \in \mathbb{R}^{N \times N \times d_{\text{model}}}$.

These features $\hat{\phi_{e}}$ are processed by $\text{MLP}_\text{edges}$ and $\text{MLP}_\text{con}$, both followed by a sigmoid function to obtain the predicted relation matrix $\hat{\phi_{e}}^c \in \mathbb{R}^{N \times N \times \mathcal{P}}$ (\ie relation categories) and the predicted connectivity matrix $\hat{\phi_{e}}^{con} \in \mathbb{R}^{N \times N \times 1}$ (\ie edge presence/absence). 
The reconstruction loss used at the decoder is: 
\begin{equation}
    \mathcal{L}_\text{deco} = 
    \lambda_\text{obj}\mathcal{L}_\text{obj} + \lambda_\text{rel}\mathcal{L}_\text{rel} + \lambda_\text{con}\mathcal{L}_\text{con},
\label{eq:deco_loss}
\end{equation}
where $\mathcal{L}_{obj}$, $\mathcal{L}_{rel}$ and $\mathcal{L}_{con}$ are the object detection, categorical relationships, and connectivity prediction losses, respectively. 
Since there is no direct correspondence between the input and decoder's reconstructed nodes, we employ the strategy from \cite{carion2020detr} to match the $N$ detected object candidates with the $M$ ground truth objects as follows:
\begin{equation}
    \mathcal{L}_\text{obj} = \Sigma_{i=1}^N [\lambda_{c}\mathcal{L}_{c}(\hat{v}_i'^c, v_i^c) + \mathbbm{1}_{v_i^c\neq\phi}(\lambda_{b}\mathcal{L}_{b}(\hat{v}_i'^b, v_i^b))],
\label{eq:object_loss}
\end{equation}
where $\mathcal{L}_{c}$ is a cross-entropy loss and the box coordinate loss $\mathcal{L}_{b}$ is the box matching cost (cf. $\mathcal{L}^b_{match}$ in \cite{carion2020detr}). 
The relation extraction loss $\mathcal{L}_\text{rel}$ and the connectivity loss $\mathcal{L}_\text{con}$ are binary cross-entropy losses computed between the predicted relation matrix $\hat{\phi_{e}}^c$ and the predicted connectivity matrix $\hat{\phi_{e}}^{con}$ against their respective ground truths.\newline

\noindent\textbf{Latent Space Regularization} 
Learning a continuous and structured latent space that effectively captures the underlying distribution of scene graphs is essential for Latent Diffusion Modeling. 
Indeed, a smooth latent space ensures that similar scene graphs map to similar latent representations, leading to consistent decoder reconstructions. 
To enforce this structure, we draw inspiration from Regularized Autoencoders (RAE)~\cite{ghosh2020regae} and optimize the following regularization term:
\begin{equation}
    \mathcal{L}_{\text{reg}} = \beta \cdot \tfrac{1}{2} {\|\mathbf{z}\|}_{2}^{2} + \lambda \cdot \|\decoparams\|^2_2,
\end{equation}
where the first term constrains latent space size to avoid unbounded optimization, and the second regularizes decoder parameters. $\beta$ and $\lambda$ are hyperparameters controlling the strength of these regularizations.

\noindent\textbf{GAE Training Objective}
Overall, the GAE is trained minimizing the loss function \(\mathcal{L}_\text{GAE} = \mathcal{L}_\text{enco} + \mathcal{L}_\text{deco} + \mathcal{L}_\text{reg}\). 
\(\mathcal{L}_\text{enco}\) ensures the latent code captures node and edge semantics, 
while \(\mathcal{L}_\text{deco}\) enforces accurate scene graph reconstruction from latent representations.  
\(\mathcal{L}_\text{reg}\) regularizes the latent space to support subsequent Latent Diffusion modeling.

\subsection{Latent Diffusion Model}
\label{subsec:stagetwo}
Diffusion models \cite{sohl2015deep, ho2020denoising} are probabilistic models that learn a data distribution $p(x)$ by iteratively denoising a variable initially sampled from a normal distribution. The working principle can be interpreted as reversing a predefined Markov chain of length $T$
through a sequence of denoising autoencoders $\model(x_{t},t),\, t=1\dots T$, each trained to predict a cleaner version of its input $x_t$ , with  $x_t$  being a noisy version of the original data $x$. 
The corresponding training objective can be simplified to:
\begin{equation}
\mathcal{L}_\text{DM} = 
\expec_{x, \epsilon \sim \mathcal{N}(0, 1), t} \Big[ \Vert \epsilon - \model(x^t, t) \Vert_{2}^{2} \Big].
\label{eq:dmloss}
\end{equation}
Here $\epsilon$ is the noise sampled from normal distribution $\mathcal{N}(0, 1)$, $\model(x^t, t)$ is the model's prediction of the noise, given the noisy data $x_t$ and the timestep $t$, $\Vert \cdot \Vert_2^2$ stands for the squared L2 norm. 

\noindent \textbf{Generative Modeling of Graphs} 
LDMs generate data efficiently by operating in a latent space, overcoming the challenges of the specific underlying data structure. 
Our Graph Encoder $\encoder$ maps input graphs into a structured latent space, encoding high-level information such as object presence, positions, appearance, and relationships. 
Unlike graphs, which are sparse and discrete, a continuous latent representation is better suited for diffusion-based models, allowing the generative process to focus on semantic content and forecast plausible developments of human activities. 
The Graph Decoder $\decoder$ then reconstructs complete scene graphs (objs + rels) from the generated latents.

\noindent \textbf{Scene Graph Anticipation through LDM} 
We train the LDM on ordered sequences of ground-truth scene graph latent vectors $\mathbf{Z}=\{\mathbf{z}_f\}_{f=0}^{F_{last}}$ obtained from our Graph Encoder $\encoder$ (as indicated in Sec.~\ref{subsec:stageone}). 
Each $\mathbf{z}_f \in \mathbb{R}^{\dimlatent}$ represents the graph latent for the $f$-th frame. $F_{last}$ is the total number of frames in the video. 
For each video, the diffusion process targets an ordered sequence of latent codes. 

The LDM is trained using fixed-size sliding windows of width $\winsize$ across the entire sequence $\mathbf{Z}$. 
At each window, we randomly select half of the tokens (each representing a temporal latent SG instance) and add Gaussian noise to these, using the remaining tokens as conditioning data to guide the denoising process. Each token in the window is positionally encoded to maintain temporal coherence. 

The denoising model $\model(\cdot, t)$ is implemented as a DiT~\cite{peebles2023dit} transformer. 
We augment the denoising model input with the diffusion-timestep embedding and token-specific embeddings to indicate whether each token $\mathbf{z}_f^t$ at diffusion-timestep $t$ is a conditioning or noised token. 
In the forward diffusion, each latent $\mathbf{z}_f^t$ is obtained by adding $t$-scheduled Gaussian noise to $\mathbf{z}_f$ from the Graph Encoder $\encoder$. 
During reverse diffusion, samples from the prior distribution $p(\mathbf{z})$ can be decoded into scene graphs in a single pass through our Graph Decoder $\decoder$. 

\subsection{Scene Graph Anticipation}
\begin{algorithm}[t]
\caption{Scene Graph Anticipation}
\label{alg:scene_graph_anticipation}
\begin{algorithmic}
\Require Observed graphs ${\graph}_{0}^{F_s}$, GAE encoder $\encoder$, GAE decoder $\decoder$, latent diffusion model $\model$, window size \winsize
\Ensure Scene graphs for future frames $\hatgraph_{F_s + 1}^{F_\text{last}}$
\State $\sequence_{\text{seen}} \gets \text{\textcolor{keywords}{encode}}({\graph}_{0}^{F_s}, \encoder)$ \Comment{\textcolor{comments}{Encode observed graphs}}
\State $\sequence_{\text{future}} \gets \text{\textcolor{keywords}{init with Gaussian noise}}$
\State $\sequence \gets \text{\textcolor{keywords}{concatenate}}(\sequence_{\text{seen}}, \sequence_{\text{future}})$ 
\State $i \gets F_s$ \Comment{\textcolor{comments}{$i$ starts at index of last observed frame}}
\While{$i < \text{\textcolor{keywords}{length}}(\sequence) - \winsize/2$}
    \State $\winsize_{\text{start}} \gets \max(0, i - \winsize/2)$ \Comment{\textcolor{comments}{Window start}}
    \State $\winsize_{\text{end}} \gets \min(\text{\textcolor{keywords}{length}}(\sequence), i + \winsize/2)$ \Comment{\textcolor{comments}{Window end}}
    \State $\sequence_{\text{window}} \gets \sequence[\winsize_{\text{start}} : \winsize_{\text{end}}]$ \Comment{\textcolor{comments}{Get current window}}
    \State $\sequence_{\text{pred}} \gets \text{\textcolor{keywords}{reverse diffusion}}(\sequence_{\text{window}}, \model)$
    \State $\sequence[i : i + \winsize/2] \gets \sequence_{\text{pred}}[\winsize/2:]$ \Comment{\textcolor{comments}{Update future}}
    \State $i \gets i + \winsize/2$ \Comment{\textcolor{comments}{Advance window}}
\EndWhile
\State $\hatgraph_{F_s + 1}^{F_\text{last}} \gets \text{\textcolor{keywords}{decode}}(\sequence_{F_s + 1}^{F_\text{last}}, \decoder)$ \Comment{\textcolor{comments}{Future SGs}}
\end{algorithmic}
\end{algorithm}
At inference, we use the observed video portion $\{I\}_{0}^{F_{s}}$ to forecast scene graphs for the future, unseen frames $\{I\}_{F_{s+1}}^{F_{last}}$. 
Following \cite{peddi2024towards}, the observed portion is represented as a sequence of graphs based solely on the visual features and category information of objects (see Sec.~\ref{subsec:stageone}). 
The Graph Encoder $\encoder$ maps the observed graphs into latent representations. Let $\mathbf{Z}_\text{seen} = {\mathbf{z}_{f=0}^{F_s}}$ denote the sequence of observed graph latents, which serve as conditioning for forecasting future latents via the LDM. 
Future latents are initialized with Gaussian noise and refined through iterative reverse diffusion, leveraging the observed temporal context. 
The process begins by centering a window of size $\winsize$ on the last observed latent $\mathbf{z}_{F_s}$. 
The first half of the window ($\winsize/2$ latents) consists of observed graph latents, while the second half contains unknown latents that have to be predicted through reverse diffusion, conditioned on the first half. 
We progressively shift the window by $\winsize/2$ after each iteration, using the previously generated latents as conditioning, and continue iterating the reverse diffusion process until the latent sequence spans the entire video length $F_\text{last}$. 
Throughout each iteration, conditioning tokens remain fixed to their initial values, as performed in \cite{wang2023pdpp}. 
This iterative approach results in a sequence of graph latents spanning the whole video. 
Finally, the Graph Decoder $\decoder$ maps the predicted latents to complete (objs + rels) scene graphs. 
The LDM anticipation procedure is shown in Alg.~\ref{alg:scene_graph_anticipation}.

\section{Experimental Setup}
\label{sec:exp}

\noindent\textbf{Dataset} 
We evaluate our approach on the Action Genome~(AG, \cite{ji2020action}) dataset, which provides detailed scene graph annotations for human-object interactions in videos. The dataset includes 35 object categories and 25 relationship categories. Relationships are classified into three types: \textit{attention} (\eg, \emph{looking at}), \textit{spatial} (\eg, \emph{in front of}), and \textit{contacting} (\eg, \emph{sitting on}). Multiple relationships can coexist between two entities. We adopt the original train and test splits from \cite{ji2020action}.

\noindent\textbf{Evaluation metrics} 
Evaluating performance for SGA is non-trivial. 
Existing literature simplifies the setting by assuming \textit{object continuity} and focusing only on the evolution of relationships (\ie edges) between pre-defined objects (\ie nodes). We argue that this choice fails to fully capture anticipation capabilities, making the task and the related evaluation metrics unable to scale to realistic use cases, including long-term activities. 
In contrast, our goal is to forecast complete scene graphs, requiring metrics to evaluate both object discovery (node forecasting) and relationship prediction (edge forecasting). 

As the first to tackle the problem of anticipating future instances of complete scene graphs, we propose a $Recall@K$ metric for Object Discovery, named \textbf{Object Recall}, with $K \in \{5, 10, 20\}$ based on the average number of objects per scene. This metric quantifies the fraction of ground-truth objects in the top-$K$ predictions. 
Object Recall measures the ability to predict relevant objects but does not penalize the prediction of object categories absent from the ground truth scene graph. 
To address this, we introduce a further metric that quantifies the similarity between the predicted set of objects $\hat{O}_f$ and the corresponding ground truth ones $O_f$. Specifically, we adopt the \textbf{Jaccard index} defined as $J_{sim}= \frac{1}{|F_{last}-F_s|}\sum_{f=F_{s+1}}^{F_{last}} \frac{|\hat{O}_f \cap O_f|}{|\hat{O}_f \cup O_f|}$. As specified by the formula, the index is computed for each future frame and averaged, with values ranging from 0 (no overlap) to 1 (perfect match)~\footnote{In AG scene graphs, each object category is annotated with a single instance — the one involved in interaction. This allows using set similarity ($\jsim$) for Object Discovery.}.

To evaluate the relationships prediction, we follow \cite{peddi2024towards} and employ the $Recall@K$ metric to evaluate predicted triplets (hereinafter \textbf{Triplets Recall}) - with $K \in \{10, 20, 50\}$. 
This metric measures the proportion of ground truth relationship triplets present in the top-$K$ predictions, providing a robust assessment of our model's accuracy in forecasting relationships. 
We report the Triplets Recall considering two working scenarios: i) \emph{With Constraint}, where for each couple of nodes we assume the existence of a single relation; ii) \emph{No Constraint}, where multiple relations are allowed between two nodes.

\noindent\textbf{Experimental setting} 
Our focus is forecasting scene graphs for future, unseen frames rather than modeling the observed video portion. With this aim, we align with \cite{peddi2024towards} and adopt their GAGS and PGAGS settings. 
In GAGS, object bounding boxes and categories for the observed video portion are sourced from AG ground truth annotations. 
In PGAGS, bounding boxes remain ground truth, while object categories are inferred using a pre-trained Faster R-CNN to reduce reliance on annotations at the observed portion.

As in \cite{peddi2024towards}, we evaluate \textbf{SGA} by varying the observed video portion and assessing forecasting at observation fractions $\mathcal{F}$=\{0.3, 0.5, 0.7, 0.9\}. 
Further, to explicitly evaluate realistic scenarios where interacted objects may change between the observed and future video portions, we introduce the \textbf{SGA under Object Distribution Shift} benchmark. 
Here, the observed portion is not taken at a fixed fraction, and an object distribution shift occurs between the last observed and future frames. For each test video, we extract up to three observed-future anticipation splits, ranked by the Jaccard distance 
\( J_\text{dist} = 1 - \frac{|O_{F_s} \cap O_{F_s+1}|}{|O_{F_s} \cup O_{F_s+1}|} \) 
where $O_{F_s}$ and $O_{F_s+1}$ are the object sets in the last observed ($F_s$) and subsequent future frame ($F_s+1$), respectively. 
Based on  $J_\text{dist}$, we categorize the anticipations splits into two difficulty levels: MID (where  $0.33 \leq \jdist < 0.66$ ) and HARD (where  $0.66 \leq \jdist \leq 1$), reflecting increasing object distribution shift. We point the reader to the supplementary material for further details on the benchmark definition.

\noindent\textbf{Implementation Details} 
Building on prior works~\cite{cong2021spatial,feng2023exploiting,peddi2024towards}, we use a Faster R-CNN~\cite{faster_rcnn} detector trained on AG~\cite{ji2020action} to extract visual features ($v^f$). In the GAGS setting, object categories ($v^c$) and bounding boxes ($v^b$) are sourced from ground truth AG annotations, while in PGAGS, $v^c$ is predicted by Faster R-CNN. Learnable semantic embeddings ($\mathbf{v}^c$) are initialized with GloVe~\cite{glove} word embeddings, as in \cite{feng2023exploiting,peddi2024towards}. 
Training progresses in two stages: first, we train the GAE ($\mathcal{L}_\text{GAE}$) and then the LDM on the frozen GAE's latent representations ($\mathcal{L}_\text{DM}$). 
At inference, the GAE encodes the observed portion graphs and decodes the LDM-forecasted latents to scene graphs. 
The GAE is trained with %
batch size of 512 for 100 epochs using the Adam optimizer with %
base learning rate of $1 \times 10^{-4}$, %
decayed via cosine annealing. The decoder $\decoder$ is configured with $N=20$ object queries. 
The LDM model uses $T=500$ diffusion timesteps with a sliding window size of $\winsize=20$. It is trained for 80k %
iterations with a batch size of 32 and Adam optimizer, starting at a learning rate of $5 \times 10^{-4}$, decayed via cosine annealing.
Further implementation details and a description of the baselines are provided in the supplementary. 

\begin{table}[t!]
    \centering
    \caption{
    SGA results in GAGS setting when varying the observed video portion ($\mathcal{F}$). 
    We use the \textcolor{gray}{gray} color for results obtained from methods assuming object continuity and predicting only the evolution of relations. 
    Top results are bold, second-best are underlined.\label{tab:anticipation_gags}}
    \vspace{-3mm}
    \resizebox{0.49\textwidth}{!}{
    \begin{tabular}{l@{~~}l@{~~}c@{~~}c@{~~}c@{~~}c@{~~}c@{~~}c@{~~}c@{~~}c@{~~}c@{~~}c@{~~}c}
            & & \multicolumn{4}{c}{\textbf{Objects}} & \multicolumn{3}{c}{\textbf{Triplets}} & \multicolumn{3}{c}{\textbf{Triplets}} \\
            & & \multicolumn{4}{c}{\textbf{Discovery}} & \multicolumn{3}{c}{%
            \textbf{With Constraint}} & \multicolumn{3}{c}{%
            \textbf{No Constraint}} \\
            \cmidrule(lr){3-6} \cmidrule(lr){7-9} \cmidrule(lr){10-12}
            $\mathcal{F}$ & \textbf{Method} & $\jsimb$ & \textbf{R@5} & \textbf{R@10} & \textbf{R@20} & \textbf{R@10} & \textbf{R@20} & \textbf{R@50} & \textbf{R@10} & \textbf{R@20} & \textbf{R@50} \\
            \midrule
            \multirow{9}{*}{0.3} 
            & STTran+ \cite{cong2021spatial} & \color{gray}{\textbf{0.69}} & - & - & - & 30.8 & 32.8 & 32.8 & 30.6 & 47.3 & 62.8\\ 
            & DSGDetr+ \cite{feng2023exploiting} & \color{gray}{\textbf{0.69}} & - & - & - & 27.0 & 28.9 & 28.9 & 30.5 & 45.1 & 62.8\\ 
            & STTran++ \cite{cong2021spatial} & \color{gray}{\textbf{0.69}} & - & - & - & 30.7 & 33.1 & 33.1 & 35.9 & 51.7 & 64.1\\ 
            & DSGDetr++ \cite{feng2023exploiting} & \color{gray}{\textbf{0.69}} & - & - & - & 25.7 & 28.2 & 28.2 & 36.1 & 50.7 & 64.0\\ 
            & SceneSayerODE \cite{peddi2024towards} & \color{gray}{\textbf{0.69}} & - & - & - & 34.9 & 37.3 & 37.3 & 40.5 & 54.1 & 63.9\\ 
            & SceneSayerSDE \cite{peddi2024towards} & \color{gray}{\textbf{0.69}} & - & - & - & 39.7 & 42.2 & 42.3 & 46.9 & 59.1 & 65.2\\ 
            \cmidrule(lr){2-12}
            & \ours ($r$=1) & 0.61 & 73.0 & 76.3 & 79.7 & 36.4 & 40.1 & 43.5 & 42.6 & 52.5 & 58.6\\
            & \ours ($r$=5) & 0.67 & \underline{77.6} & \underline{80.5} & \underline{83.2} & \underline{42.7} & \underline{46.4} & \underline{49.7} & \underline{50.6} & \underline{60.4} & \underline{66.2}\\
            & \ours ($r$=10) & \underline{0.68} & \textbf{78.7} & \textbf{81.4} & \textbf{83.8} & \textbf{44.3} & \textbf{47.9} & \textbf{51.2} & \textbf{52.6} & \textbf{62.2} & \textbf{67.8}\\
            \hline
            \hline
            \\ [-0.8em]
            \multirow{9}{*}{0.5} 
            & STTran+ \cite{cong2021spatial} & \color{gray}{\textbf{0.73}} & - & - & - & 35.0 & 37.1 & 37.1 & 34.4 & 53.4 & 70.8\\ 
            & DSGDetr+ \cite{feng2023exploiting} & \color{gray}{\textbf{0.73}} & - & - & - & 31.2 & 33.3 & 33.3 & 34.3 & 51.0 & 70.8\\ 
            & STTran++ \cite{cong2021spatial} & \color{gray}{\textbf{0.73}} & - & - & - & 35.6 & 38.1 & 38.1 & 40.3 & 58.4 & 72.2\\ 
            & DSGDetr++ \cite{feng2023exploiting} & \color{gray}{\textbf{0.73}} & - & - & - & 29.3 & 31.9 & 32.0 & 40.3 & 56.9 & 72.0\\ 
            & SceneSayerODE \cite{peddi2024towards} & \color{gray}{\textbf{0.73}} & - & - & - & 40.7 & 43.4 & 43.4 & 47.0 & 62.2 & 72.4\\ 
            & SceneSayerSDE \cite{peddi2024towards} & \color{gray}{\textbf{0.73}} & - & - & - & 45.0 & 47.7 & 47.7 & 52.5 & 66.4 & \underline{73.5}\\ 
            \cmidrule(lr){2-12}
            & \ours ($r$=1) & 0.65 & 77.6 & 81.0 & 83.7 & 40.3 & 44.4 & 47.9 & 46.8 & 58.2 & 65.3\\
            & \ours ($r$=5) & 0.70 & \underline{82.5} & \underline{85.2} & \underline{87.3} & \underline{47.3} & \underline{51.2} & \underline{54.6} & \underline{55.6} & \underline{66.6} & 73.3\\
            & \ours ($r$=10) & \underline{0.72} & \textbf{83.5} & \textbf{86.1} & \textbf{88.0} & \textbf{49.1} & \textbf{52.9} & \textbf{56.2} & \textbf{58.0} & \textbf{68.5} & \textbf{74.9}\\
            \hline
            \hline
            \\ [-0.8em]
            \multirow{9}{*}{0.7} 
            & STTran+ \cite{cong2021spatial} & \color{gray}{\textbf{0.81}} & - & - & - &  40.0 & 41.8 & 41.8 & 41.0 & 62.5 & 80.4 \\ 
            & DSGDetr+ \cite{feng2023exploiting} & \color{gray}{\textbf{0.81}} & - & - & - & 35.5 & 37.3 & 37.3 & 41.0 & 59.8 & 80.7 \\ 
            & STTran++ \cite{cong2021spatial} & \color{gray}{\textbf{0.81}} & - & - & - & 41.3 & 43.6 & 43.6 & 48.2 & 68.8 & 82.0 \\ 
            & DSGDetr++ \cite{feng2023exploiting} & \color{gray}{\textbf{0.81}} & - & - & - & 33.9 & 36.3 & 36.3 & 48.0 & 66.7 & 81.9\\ 
            & SceneSayerODE \cite{peddi2024towards} & \color{gray}{\textbf{0.81}} & - & - & - & 49.1 & 51.6 & 51.6 & 58.0 & 74.0 & 82.8\\ 
            & SceneSayerSDE \cite{peddi2024towards} & \color{gray}{\textbf{0.81}} & - & - & - & 52.0 & 54.5 & 54.5 & 61.8 & \underline{76.7} & \underline{83.4}\\ 
            \cmidrule(lr){2-12}
            & \ours ($r$=1) & 0.72 & 84.1 & 86.9 & 89.0 & 47.9 & 52.0 & 55.5 & 55.5 & 68.0 & 75.2\\
            & \ours ($r$=5)  & 0.78 & \underline{88.5} & \underline{90.5} & \underline{92.0} & \underline{55.0} & \underline{58.7} & \underline{62.2} & \underline{65.2} & 76.3 & 82.4\\
            & \ours ($r$=10) & \underline{0.79} & \textbf{89.4} & \textbf{91.4} & \textbf{92.7} & \textbf{56.7} & \textbf{60.4} & \textbf{63.8} & \textbf{67.7} & \textbf{78.1} & \textbf{84.0}\\
            \hline
            \hline
            \\ [-0.8em]
            \multirow{9}{*}{0.9} 
            & STTran+ \cite{cong2021spatial} & \color{gray}{\textbf{0.89}} & - & - & - & 44.7 & 45.9 & 45.9 & 50.9 & 74.8 & 90.9\\ 
            & DSGDetr+ \cite{feng2023exploiting} & \color{gray}{\textbf{0.89}} & - & - & - & 38.8 & 40.0 & 40.0 & 51.0 & 71.7 & 91.2\\ 
            & STTran++ \cite{cong2021spatial} & \color{gray}{\textbf{0.89}} & - & - & - & 46.0 & 47.7 & 47.7 & 60.2 & 81.5 & 92.3\\ 
            & DSGDetr++ \cite{feng2023exploiting} & \color{gray}{\textbf{0.89}} & - & - & - & 38.1 & 39.8 & 39.8 & 58.8 & 78.8 & 92.2 \\ 
            & SceneSayerODE \cite{peddi2024towards} & \color{gray}{\textbf{0.89}} & - & - & - & 58.1 & 59.8 & 59.8 & 72.6 & 86.7 & \underline{93.2}\\ 
            & SceneSayerSDE \cite{peddi2024towards} & \color{gray}{\textbf{0.89}} & - & - & - & 60.3 & 61.9 & 61.9 & 74.8 & \underline{88.0} & \textbf{93.5}\\ 
            \cmidrule(lr){2-12}
            & \ours ($r$=1) & 0.82 & 91.3 & 93.2 & 94.3 & 55.7 & 59.6 & 63.8 & 67.4 & 80.1 & 86.5\\
            & \ours ($r$=5) & \underline{0.87} & \underline{94.7} & \underline{96.0} & \underline{96.8} & \underline{63.4} & \underline{66.9} & \underline{70.7} & \underline{77.5} & 87.0 & 92.2\\
            & \ours ($r$=10) & \underline{0.87} & \textbf{95.1} & \textbf{96.5} & \textbf{97.1} & \textbf{64.5} & \textbf{68.0} & \textbf{71.7} & \textbf{79.7} & \textbf{88.1} & 92.9\\
            \hline
        \end{tabular}
    }
     \vspace{-2mm}
\end{table}

\begin{table}[t]
    \centering
    \caption{
    SGA results in PGAGS setting when varying the observed video portion ($\mathcal{F}$). Values in \textcolor{gray}{gray} assume object continuity. 
    Top results are bold, second-best are underlined.
    \label{tab:anticipation_pgags}}
    \vspace{-3mm}
    \resizebox{0.49\textwidth}{!}{
    \begin{tabular}{l@{~~}l@{~~}c@{~~}c@{~~}c@{~~}c@{~~}c@{~~}c@{~~}c@{~~}c@{~~}c@{~~}c@{~~}c}
            & & \multicolumn{4}{c}{\textbf{Objects}} & \multicolumn{3}{c}{\textbf{Triplets}} & \multicolumn{3}{c}{\textbf{Triplets}} \\
            & & \multicolumn{4}{c}{\textbf{Discovery}} & \multicolumn{3}{c}{%
            \textbf{With Constraint}} & \multicolumn{3}{c}{%
            \textbf{No Constraint}} \\
            \cmidrule(lr){3-6} \cmidrule(lr){7-9} \cmidrule(lr){10-12}
            $\mathcal{F}$ & \textbf{Method} & $\jsimb$ & \textbf{R@5} & \textbf{R@10} & \textbf{R@20} & \textbf{R@10} & \textbf{R@20} & \textbf{R@50} & \textbf{R@10} & \textbf{R@20} & \textbf{R@50} \\
            \midrule
            \multirow{5}{*}{0.3} 
            & SceneSayerODE \cite{peddi2024towards} & \color{gray}\underline{0.55} & - & - & - & 27.0 & 27.9 & 27.9 & 33.0 & 40.9 & 46.5\\
            & SceneSayerSDE \cite{peddi2024towards} & \color{gray}\underline{0.55} & - & - & - & 28.8 & 29.9 & 29.9 & 34.6 & 42.0 & 46.2\\
            \cmidrule(lr){2-12}
            & \ours ($r$=1) & 0.48 & 62.1 & 66.8 & 71.7 & 24.7 & 27.6 & 31.6 & 29.9 & 34.8 & 39.1\\
            & \ours ($r$=5) & \underline{0.55} & \underline{68.0} & \underline{71.9} & \underline{76.0} & \underline{31.6} & \underline{34.5} & \underline{38.5} & \underline{39.4} & \underline{44.7} & \underline{48.7}\\
            & \ours ($r$=10) & \textbf{0.57} & \textbf{69.8} & \textbf{73.5} & \textbf{77.4} & \textbf{33.9} & \textbf{36.8} & \textbf{40.8} & \textbf{42.4} & \textbf{47.8} & \textbf{51.8}\\
            \hline
            \hline
            \\ [-0.8em]
            \multirow{5}{*}{0.5} 
            & SceneSayerODE \cite{peddi2024towards} & \color{gray}\underline{0.58} & - & - & - & 30.5 & 31.5 & 31.5 & 36.8 & 45.9 & 51.8\\
            & SceneSayerSDE \cite{peddi2024towards} & \color{gray}\underline{0.58} & - & - & - & 32.2 & 33.3 & 33.3 & 38.4 & 46.9 & 51.8\\
            \cmidrule(lr){2-12}
            & \ours ($r$=1) & 0.51 & 65.6 & 69.5 & 74.1 & 28.0 & 30.9 & 34.6 & 33.6 & 39.6 & 44.2\\
            & \ours ($r$=5) & 0.57 & \underline{71.2} & \underline{74.9} & \underline{78.4} & \underline{34.9} & \underline{37.9} & \underline{41.6} & \underline{42.9} & \underline{48.9} & \underline{53.6}\\
            & \ours ($r$=10) & \textbf{0.60} & \textbf{72.9} & \textbf{76.3} & \textbf{79.8} & \textbf{37.0} & \textbf{40.0} & \textbf{43.9} & \textbf{46.1} & \textbf{52.2} & \textbf{56.5}\\
            \hline
            \hline
            \\ [-0.8em]
            \multirow{5}{*}{0.7} 
            & SceneSayerODE \cite{peddi2024towards} & \color{gray}\underline{0.63} & - & - & - & 36.5 & 37.3 & 37.3 & 44.6 & 54.4 & 60.3\\
            & SceneSayerSDE \cite{peddi2024towards} & \color{gray}\underline{0.63} & - & - & - & 37.6 & 38.5 & 38.5 & 45.6 & 54.6 & 59.3\\
            \cmidrule(lr){2-12}
            & \ours ($r$=1) & 0.55 & 69.5 & 73.7 & 77.6 & 32.1 & 35.1 & 39.3 & 38.5 & 45.2 & 50.0\\
            & \ours ($r$=5) & 0.62 & \underline{75.3} & \underline{78.7} & \underline{81.9} & \underline{39.7} & \underline{42.7} & \underline{46.7} & \underline{49.4} & \underline{55.4} & \underline{60.1}\\
            & \ours ($r$=10) & \textbf{0.64} & \textbf{77.1} & \textbf{80.4} & \textbf{83.2} & \textbf{41.9} & \textbf{45.0} & \textbf{49.1} & \textbf{52.9} & \textbf{58.}8 & \textbf{63.0}\\
            \hline
            \hline
            \\ [-0.8em]
            \multirow{5}{*}{0.9} 
            & SceneSayerODE \cite{peddi2024towards} & \color{gray}\textbf{0.70} & - & - & - & 41.6 & 42.2 & 42.2 & 52.7 & 61.8 & 66.5\\
            & SceneSayerSDE \cite{peddi2024towards} & \color{gray}\textbf{0.70} & - & - & - & 42.5 & 43.1 & 43.1 & 53.8 & 62.4 & 66.2\\
            \cmidrule(lr){2-12}
            & \ours ($r$=1) & 0.62 & 75.2 & 78.8 & 82.1 & 38.2 & 41.5 & 46.0 & 46.6 & 52.8 & 58.2\\
            & \ours ($r$=5) & \underline{0.68} & \underline{80.4} & \underline{83.4} & \underline{86.1} & \underline{46.1} & \underline{49.2} & \underline{53.7} & \underline{58.2} & \underline{63.3} & \underline{67.6}\\
            & \ours ($r$=10) & \textbf{0.70} & \textbf{81.9} & \textbf{84.6} & \textbf{87.2} & \textbf{48.4} & \textbf{51.4} & \textbf{55.9} & \textbf{61.4} & \textbf{66.2} & \textbf{70.2}\\
            \hline
        \end{tabular}
    }
    \vspace{-4mm}
\end{table}

\section{Results}

\noindent\textbf{Scene Graph Anticipation}
We evaluate SGA in terms of Object Discovery and Triplet Prediction.
Unlike competitors assuming object continuity, \ours can predict both appearing and disappearing objects as well as changes in their relationships. 
Moreover, thanks to its generative nature, \ours is the only method able to predict multiple plausible future activity developments, all evolving from the same partial video observation by simply varying the diffusion seed. 
Thus, we report results when considering one single prediction ($r$=1) as well as several versions of future SGs sequences ($r$=5 and $r$=10). In the latter case, we keep the best output over $r$ predicted sequences generated by our model in terms of R@10 Triplets No Constraint. 

The GAGS setting results in Tab.~\ref{tab:anticipation_gags} 
indicate that relying on object continuity across frames may provide a small advantage in terms of Object Discovery but largely hinders Triplet Prediction. 
\ours thrives by forecasting complete scene graphs, a more challenging task than predicting the evolution of relations between fixed nodes. This capability is essential  for modeling the temporal dynamics of daily human activities. 
Notably, while \ours does not enforce object continuity, it preserves nodes when changes are unlikely, as reflected in its high Object Recall when 90\% of the video is observed ($\mathcal{F}=0.9$). 

In Tab.~\ref{tab:anticipation_pgags}, we compare \ours with the top two competitors (SceneSayerODE, SceneSayerSDE \cite{peddi2024towards}) in the PGAGS setting, which relaxes the reliance on object annotations for modeling the observed video portion. 
Here, competitors still rely on object continuity. However, they must label the objects in last observed frame (as the object categories are not sourced from AG annotations) besides predicting how their relationship evolve over time. 
In this more challenging scenario, \ours outperforms the competitors in both Object Discovery and Triplet Prediction.

\begin{table}[t]%
    \centering
    \caption{    
    SGA under Object Distribution Shift results in GAGS (top) and PGAGS (bottom) setting. 
    Values in \textcolor{gray}{gray} assume object continuity. 
    \label{tab:subset_gags}}
    \vspace{-4mm}
    \resizebox{0.48\textwidth}{!}{
    \begin{tabular}{l@{~~}l@{~~}r@{~~}r@{~~}r@{~~}r@{~~}r@{~~}r@{~~}r@{~~}r@{~~}r@{~~}r}
        & \multicolumn{4}{c}{\textbf{Objects}} & \multicolumn{3}{c}{\textbf{Triplets}} & \multicolumn{3}{c}{\textbf{Triplets}} \\
        & \multicolumn{4}{c}{\textbf{Discovery}} & \multicolumn{3}{c}{\textbf{With Constraint}} & \multicolumn{3}{c}{\textbf{No Constraint}} \\
        \cmidrule(lr){2-5} \cmidrule(lr){6-8} \cmidrule(lr){9-11}
        \textbf{Method} & $\jsimb$ & \textbf{R@5} & \textbf{R@10} & \textbf{R@20} & \textbf{R@10} & \textbf{R@20} & \textbf{R@50} & \textbf{R@10} & \textbf{R@20} & \textbf{R@50} \\
        \midrule
        \rowcolor{red!20}\multicolumn{11}{c}{\textbf{MID difficulty scenario - GAGS}}\\
        SceneSayerODE \cite{peddi2024towards} & \color{gray}{0.60} & - & - & - & 36.1 & 37.7 & 37.7 & 44.7 & 56.7 & 63.5 \\
        SceneSayerSDE \cite{peddi2024towards} & \color{gray}{0.60} & - & - & - & 39.3 & 41.2 & 41.2 & 49.0 & \underline{59.6} & 64.1 \\
        \ours ($r$=1) & 0.54 & 71.0 & 74.3 & 77.7 & 33.7 & 37.2 & 40.7 & 39.4 & 49.1 & 55.2 \\
        \ours ($r$=5) & \underline{0.61} & \underline{76.9} & \underline{79.4} & \underline{82.0} & \underline{41.6} & \underline{44.9} & \underline{48.5} & \underline{49.8} & 59.1 & \underline{64.8} \\
        \ours ($r$=10) & \textbf{0.62} & \textbf{78.1} & \textbf{80.6} & \textbf{83.0} & \textbf{43.5} & \textbf{46.8} & \textbf{50.4} & \textbf{52.6} & \textbf{61.5} & \textbf{66.7} \\
        \rowcolor{red!40}\multicolumn{11}{c}{\textbf{HARD difficulty scenario - GAGS}}\\
        SceneSayerODE \cite{peddi2024towards} & \color{gray}{0.40} & - & - & - & 14.3 & 14.6 & 14.6 & 18.9 & 22.7 & 24.7 \\
        SceneSayerSDE \cite{peddi2024towards} & \color{gray}{0.40} & - & - & - & 15.4 & 15.8 & 15.8 & 20.3 & 23.7 & 24.9 \\
        \ours ($r$=1) & 0.38 & 56 & 58.9 & 63.9 & 15.5 & 17.5 & 20.7 & 17.9 & 21.4 & 24.5 \\
        \ours ($r$=5) & \underline{0.43} & \underline{61.0} & \underline{63.9} & \underline{68.3} & \underline{21.6} & \underline{23.9} & \underline{27.2} & \underline{25.5} & \underline{29.8} & \underline{33.4} \\
        \ours ($r$=10) & \textbf{0.45} & \textbf{63.1} & \textbf{65.8} & \textbf{69.9} & \textbf{24.1} & \textbf{26.3} & \textbf{29.5} & \textbf{28.8} & \textbf{33.4} & \textbf{36.9}\\
        \midrule
        \rowcolor{red!20}\multicolumn{11}{c}{\textbf{MID difficulty scenario - PGAGS}}\\
        SceneSayerODE \cite{peddi2024towards} & \color{gray}0.51 & - & - & - & 27.3 & 28.2 & 28.2 & 35.7 & 43.4 & 47.6\\
        SceneSayerSDE \cite{peddi2024towards} & \color{gray}0.51 & - & - & - & 28.4 & 29.4 & 29.4 & 36.5 & 43.6 & 47.2\\
        \ours ($r$=1) & 0.45 & 61.4 & 65.7 & 70.7 & 24.1 & 26.8 & 30.7 & 29.1 & 34.1 & 38.2\\
        \ours ($r$=5) & \underline{0.53} & \underline{68.0} & \underline{71.8} & \underline{75.8} & \underline{31.8} & \underline{34.6} & \underline{38.6} & \underline{39.8} & \underline{45.2} & \underline{49.2}\\
        \ours ($r$=10) & \textbf{0.55} & \textbf{70.1} & \textbf{73.6} & \textbf{77.3} & \textbf{34.2} & \textbf{37.0} & \textbf{41.0} & \textbf{43.3} & \textbf{48.8} & \textbf{52.8}\\
        \rowcolor{red!40}\multicolumn{11}{c}{\textbf{HARD difficulty scenario - PGAGS}}\\
        SceneSayerODE \cite{peddi2024towards} & \color{gray}0.37 & - & - & - & 10.3 & 10.6 & 10.6 & 13.9 & 16.4 & 17.5\\
        SceneSayerSDE \cite{peddi2024towards} & \color{gray}0.37 & - & - & - & 10.7 & 11.1 & 11.1 & 14.9 & 17.1 & 18.0\\
        \ours ($r$=1) & 0.36 & 52.8 & 56.8 & 62.2 & 11.3 & 13.4 & 16.8 & 12.8 & 14.8 & 18.3\\
        \ours ($r$=5) & \underline{0.40} & \underline{57.7} & \underline{61.4} & \underline{66.2} & \underline{16.8} & \underline{19.1} & \underline{22.8} & \underline{20.1} & \underline{22.7} & \underline{26.5}\\
        \ours ($r$=10) & \textbf{0.43} & \textbf{60.5} & \textbf{64.1} & \textbf{68.5} & \textbf{19.8} & \textbf{22.0} & \textbf{25.7} & \textbf{24.2} & \textbf{27.5} & \textbf{31.3}\\
        \bottomrule
    \end{tabular}
    }
    \vspace{-4mm}
\end{table}

\noindent\textbf{SGA under Object Distribution Shift} 
We assess the performance of top SGA methods in our novel benchmark. 

We report GAGS results in the top part of Tab.~\ref{tab:subset_gags}. 
In the MID scenario, where the object distribution shift is moderate, SceneSayer’s object continuity assumption may hold, leading to competitive Triplet performance. 
However, \ours surpasses SceneSayer from $r=5$ onward, achieving higher Object Discovery and Triplet scores. 
In the HARD scenario, where the object distribution shift is more severe, SceneSayer’s reliance on object continuity fails, resulting in a significant drop in $J_{sim}$ and Triplet performance. 
In contrast, \ours excels at object forecasting, achieving superior $J_{sim}$ and significantly improving Triplet predictions.

A similar trend is observed for the PGAGS results in the bottom part of Tab.~\ref{tab:subset_gags}, here the advantage of \ours over the best competitors is even more pronounced.

\subsection {Design Choices and Ablation}
\noindent\textbf{LDM timesteps}
\ours employs $T$=500 diffusion timesteps for the LDM. In Tab.~\ref{tab:diff_win_ablations} (top), we also evaluate $T$=\{200, 1000\}. Notably, $T$=500 is sufficient for state-of-the-art results, while higher values enhance performance at the cost of increased computational complexity. 

\noindent\textbf{Window size} 
\ours applies the reverse diffusion process iteratively using a sliding window of size $\winsize$=20. The window size is crucial as it determines the available context at each step. 
We evaluate $\winsize \in \{10, 20, 40, \text{Whole video}\}$, with results in Tab.~\ref{tab:diff_win_ablations} (middle). Larger windows (40 or $\text{Whole video}$) slightly degrade performance, indicating that excessive context can be detrimental. 
While $\winsize$=10 yields the best results, $\winsize$=20 strikes a balance between performance and computational efficiency, as smaller windows require more sliding steps (see Alg.\ref{alg:scene_graph_anticipation}).

\noindent\textbf{Encoder Auxiliary Losses}
We ablate the adoption of the encoder auxiliary losses $\mathcal{L}_{\text{nodes}}$ and $\mathcal{L}_{\text{edges}}$. 
By classifying nodes and edges, these losses encourage features to retain essential semantic information before the max-pooling operation that outputs the global graph latent code $\mathbf{z}$. 
The results in Tab.~\ref{tab:diff_win_ablations} (bottom)  show that incorporating auxiliary encoder losses slightly enhances our model’s performance. These findings suggest that auxiliary encoder losses help enrich the graph latent representation $\mathbf{z}$ with useful information, thereby reducing the decoder’s burden in accurately predicting the correct nodes and edges.

\begin{table}[t]
    \centering
    \caption{SGA in GAGS setting results when varying the number of diffusion steps $T$ (top), the diffusion window size $S$ (middle), and ablating the auxiliary losses (bottom). Results are reported for $\mathcal{F}=0.3$ and $r$=10.\label{tab:diff_win_ablations}}
    \vspace{-3mm}
    \resizebox{0.48\textwidth}{!}{
    \begin{tabular}{c@{~~~}c@{~~~}c@{~~~}c@{~~~}c@{~~~}c@{~~~}c@{~~~}c@{~~~}c@{~~~}c@{~~~}c@{~~~}c@{~~~}c}
        & & & \multicolumn{4}{c}{\textbf{Objects}} & \multicolumn{3}{c}{\textbf{Triplets}} & \multicolumn{3}{c}{\textbf{Triplets}} \\
        & & \multirow{2}{*}{\makecell{\textbf{Aux.}\\\textbf{Loss}}} & \multicolumn{4}{c}{\textbf{Discovery}} & \multicolumn{3}{c}{\textbf{With Constraint}} & \multicolumn{3}{c}{\textbf{No Constraint}} \\
        \cmidrule(lr){4-7} \cmidrule(lr){8-10} \cmidrule(lr){11-13}
        \textbf{T} & \textbf{\winsize} & & $\jsimb$ & \textbf{R@5} & \textbf{R@10} & \textbf{R@20} & \textbf{R@10} & \textbf{R@20} & \textbf{R@50} & \textbf{R@10} & \textbf{R@20} & \textbf{R@50} \\
        \midrule
        200 & 20 & \cmark & 0.66 & 77.7 & 80.4 & 83.1 & 43.3 & 46.8 & 50.2 & 51.7 & 61.1 & 66.1\\
        500 & 20 & \cmark & \textbf{0.68} & \underline{78.7} & \underline{81.4} & \underline{83.8} & \underline{44.3} & \underline{47.9} & \underline{51.2} & \underline{52.6} & \underline{62.2} & \underline{67.8}\\
        1000 & 20 & \cmark & \underline{0.67} & \textbf{79.3} & \textbf{82.1} & \textbf{84.4} & \textbf{45.0} & \textbf{48.7} & \textbf{51.9} & \textbf{53.1} & \textbf{63.1} & \textbf{68.8}\\
        \midrule
        500 & 10 & \cmark & \textbf{0.68} & \textbf{79.0} & \textbf{81.5} & \textbf{83.8} & \textbf{45.1} & \textbf{48.3} & \textbf{52.0} & \textbf{53.8} & \textbf{63.1} & \textbf{68.1}\\
        500 & 20 & \cmark & \textbf{0.68} & \underline{78.7} & \underline{81.4} & \textbf{83.8} & \underline{44.3} & \underline{47.9} & \underline{51.2} & \underline{52.6} & \underline{62.2} & \underline{67.8}\\
        500 & 40 & \cmark & \underline{0.65} & 76.3 & 79.5 & \underline{82.1} & 41.5 & 45.3 & 48.5 & 49.2 & 58.2 & 64.0\\
        500 & \text{Whole video} & \cmark & 0.60 & 71.9 & 75.0 & 77.6 & 36.2 & 39.4 & 42.2 & 43.4 & 51.0 & 56.8\\
        \midrule
        500 & 20 & 
        \xmark & \textbf{0.68} & 78.5 & \textbf{81.8} & \textbf{85.2} & 43.6 & 46.8 & 49.8 & \textbf{53.0} & 62.0 & 66.9\\
         500 & 20 &
        \cmark & \textbf{0.68} & \textbf{78.7} & 81.4 & 83.8 & \textbf{44.3} & \textbf{47.9} & \textbf{51.2} & 52.6 & \textbf{62.2} & \textbf{67.8}\\
        
    \bottomrule
    \end{tabular}
    }
    \vspace{-4mm}
\end{table}

\section{Conclusion}
\label{sec:conclusion}
In this paper we presented \ours, the first method for Scene Graph Anticipation that can handle dynamic scenarios where objects may appear and disappear over time. This property is essential for modeling human-environment interactions in unstructured daily activity videos through scene graph sequences.

\ours leverages a Graph Auto-Encoder to map the observed video portion into latent vectors, and a conditional Latent Diffusion Model to predict complete future scene graphs (\ie objects and their relationship) based on the observed temporal context. 

A thorough experimental analysis, including evaluations on a newly introduced benchmark with settings of increasing difficulty, showed that \ours achieves top results in Object Discovery and Triplets Eval while tackling a significantly harder task than its competitors, which focus solely on the evolution of relationships between fixed objects. 
We will release our code to foster further research in modeling human activities in videos.
{
    \small
    \bibliographystyle{ieeenat_fullname}
    \bibliography{main}
}

\clearpage
\section*{Appendix}
\setcounter{section}{0}
\renewcommand{\thesection}{\Alph{section}}

\section{Limitations and Future Work}
The primary strength of \ours lies in its ability to predict complete scene graphs, modeling both object and relationship evolution in human daily activity videos. However, this approach complicates direct comparisons with prior methods assuming object continuity, as relationship prediction in \ours is intrinsically linked to object forecasting.
To ensure a comprehensive evaluation, we complement Relation Discovery metrics with Object Discovery metrics and introduce a novel benchmark for assessing SGA under object distribution shift. 
Further progress could involve exploring additional evaluation strategies, potentially including objects' spatial localization.

\ours represents a significant advancement in modeling human activities by eliminating the assumption of object continuity when forecasting scene graphs. Future work should further push the boundaries of SGA by going beyond closed-set object and relationship categories to address the complexity and variability of real-world human-environment interactions.

Lastly, we explored diffusion generative models for modeling the temporal evolution of scene graph latents. As a flexible framework, \ours can seamlessly integrate various generative models (\eg, autoregressive, flow-based), or combine our diffusion model with efficient sampling techniques to enhance performance and reduce the computational complexity of DDPM’s iterative generation. This is a direction we plan to explore in the future.

\section{Design Choices Validation}
This section presents ablation studies validating the technical choices of \ours, focusing on SGA in the GAGS~\cite{peddi2024towards} setting.

\subsection{Number of Object Queries} 
The number of object queries (or query slots) sets an upper bound on the maximum number of object proposals in a single scene graph. 
Inspired by DETR ~\cite{carion2020detr}, we set the number of object queries in the Graph Decoder (\(\decoder\)) to $N=20$, exceeding the maximum number of object instances in Action Genome scene graphs (\ie, 10). 
In Tab.~\ref{tab:ablation_nqueries}, we compare our default $N=20$ with $N=10$, showing comparable results. 

\subsection{Latent Space Regularization}
\label{subsec:latent_reg}
A common approach to obtaining a smooth and structured latent space for latent diffusion models (LDMs) is to use a Variational Autoencoder (VAE), where a reconstruction loss is optimized jointly with a regularization term to enforce a fixed prior distribution on the latent space. In standard VAEs~\cite{higgins2017beta}, this regularization is given by the Kullback-Leibler (KL) divergence:
\begin{equation}
\mathcal{L}_{\text{reg}}^{\text{VAE}} = \beta_{\text{VAE}} \mathcal{L}_{\text{KL}}~.
\end{equation}
However, improper tuning of $\beta_{\text{VAE}}$ can lead to poor reconstructions due to over-regularization.

To avoid these issues, we adopt a Regularized Autoencoder (RAE, \cite{ghosh2020regae}), which replaces the KL regularization term with: 
\begin{equation}
    \mathcal{L}_{\text{reg}}^{\text{RAE}} = \beta \tfrac{1}{2} {\|\mathbf{z}\|}_{2}^{2} + \lambda \|\decoparams\|^2_2~,
\end{equation}
where $\|\mathbf{z}\|_2^2$ prevents unbounded optimization of the latent space, and $\|\decoparams\|_2^2$ regularizes the decoder parameters.

We compare our default RAE choice to a $\beta$-VAE, setting $\beta_{\text{VAE}} = 0.1$ based on empirical tuning, as larger values significantly degraded the reconstruction quality of the VAE (first stage of training). 
Scene Graph Anticipation results in Tab.~\ref{tab:suppl_vae} show that when $\beta_{\text{VAE}}$ is properly tuned, the VAE performs comparably to RAE. 
However, the RAE is easier to train -- due to the absence of KL divergence -- while providing a well-structured latent space, making it a natural choice for our LDM framework. 

\begin{table}[!tb]
    \centering
    \caption{
    SGA results in GAGS setting of \ours when varying the number of object queries $N$. Results are reported for $\mathcal{F}$=0.3 and $r$=10.~\label{tab:ablation_nqueries}}
    \resizebox{0.48\textwidth}{!}{
    \begin{tabular}{c@{~~~}c@{~~~}c@{~~~}c@{~~~}c@{~~~}c@{~~~}c@{~~~}c@{~~~}c@{~~~}c@{~~~}c}
        & \multicolumn{4}{c}{\textbf{Objects}} & \multicolumn{3}{c}{\textbf{Triplets}} & \multicolumn{3}{c}{\textbf{Triplets}} \\
        & \multicolumn{4}{c}{\textbf{Discovery}} & \multicolumn{3}{c}{\textbf{With Constraint}} & \multicolumn{3}{c}{\textbf{No Constraint}} \\
        \cmidrule(lr){3-5} \cmidrule(lr){6-8} \cmidrule(lr){9-11}
        \textbf{$N$} & $\jsimb$ & \textbf{R@5} & \textbf{R@10} & \textbf{R@20} & \textbf{R@10} & \textbf{R@20} & \textbf{R@50} & \textbf{R@10} & \textbf{R@20} & \textbf{R@50} \\
        \midrule
        10 & \textbf{0.68} & \textbf{79.4} & \textbf{82.0} & 82.0 & 43.8 & 46.6 & 48.5 & \textbf{52.7} & \textbf{62.8} & \textbf{67.8}\\
        20 & \textbf{0.68} & 78.7 & 81.4 & \textbf{83.8} & \textbf{44.3} & \textbf{47.9} & \textbf{51.2} & 52.6 & 62.2 & \textbf{67.8}\\
        \hline
    \end{tabular}
    }
\end{table}

\begin{table}[t]
    \centering
    \caption{SGA results in GAGS setting of \ours with two scene graph embeddings: one from a Variational Autoencoder (VAE) and the other, our default, from a Regularized Autoencoder (RAE). Results are reported for $\mathcal{F}$=0.3 and $r$=10.\label{tab:suppl_vae}}
    \resizebox{0.48\textwidth}{!}{
    \begin{tabular}{c@{~~~}c@{~~~}c@{~~~}c@{~~~}c@{~~~}c@{~~~}c@{~~~}c@{~~~}c@{~~~}c@{~~~}c}
        & \multicolumn{4}{c}{\textbf{Objects}} & \multicolumn{3}{c}{\textbf{Triplets}} & \multicolumn{3}{c}{\textbf{Triplets}} \\
        & \multicolumn{4}{c}{\textbf{Discovery}} & \multicolumn{3}{c}{\textbf{With Constraint}} & \multicolumn{3}{c}{\textbf{No Constraint}} \\
        \cmidrule(lr){3-5} \cmidrule(lr){6-8} \cmidrule(lr){9-11}
        \textbf{SG emb.} & $\jsimb$ & \textbf{R@5} & \textbf{R@10} & \textbf{R@20} & \textbf{R@10} & \textbf{R@20} & \textbf{R@50} & \textbf{R@10} & \textbf{R@20} & \textbf{R@50}\\
        \midrule
        \text{VAE} & 0.65 & \textbf{79.3} & \textbf{83.1} & \textbf{86.2} & 44.0 & \textbf{48.6} & \textbf{54.3} & \textbf{53.1} & \textbf{63.3} & \textbf{69.4}\\
        \text{RAE} & \textbf{0.68} & 78.7 & 81.4 & 83.8 & \textbf{44.3} & 47.9 & 51.2 & 52.6 & 62.2 & 67.8\\
    \bottomrule
    \end{tabular}
    }
\end{table}

\section{Implementation Details}
In this section, we provide additional technical details related to the architectures employed and the corresponding hyperparameter selection. 

\smallskip
\noindent\textbf{Graph Encoder} For the encoder $\encoder$, we used the GCNN architecture proposed by \cite{johnson2018image}.  
The encoder consists of five stacked GCNN layers followed by two MLP heads tailored for node and edge classification tasks. Specifically, the MLP heads receive node features $\phi_{v}$ with dimensions $(|\mathcal{V}| \times \dimnode)$ and edge features $\phi_{e}$ with dimensions $(|\mathcal{E}| \times \dimedge)$ from the last GCNN layer. 
The output $\hat{v}^c$ is provided with dimensions $(|\mathcal{V}| \times \mathcal{C})$ and $\hat{p}^c$ with dimensions $(|\mathcal{E}| \times \mathcal{P})$, representing predictions over the categorical distributions of nodes and edges, respectively\footnote{In our notation \textit{hat} indicates model prediction.}. 
These predictions are used to compute the auxiliary losses $\mathcal{L}_{\text{nodes}}$ and $\mathcal{L}_{\text{edges}}$ in Eq. 4 %
of the main paper:

\begin{align}
\mathcal{L}_{\text{nodes}} &= -\frac{1}{\mathcal{V}} \sum_{i=0}^{|\mathcal{V}|} \sum_{c=0}^{\mathcal{C}} v_i^c \log(\hat{v}_i^c) \label{eq:node_loss} \\
\begin{split}
    \mathcal{L}_{\text{edges}} &= -\frac{1}{\mathcal{E}} \sum_{j=0}^{|\mathcal{E}|} \sum_{c=0}^{\mathcal{P}} \bigg[ p^c_j \log\big(\sigma(\hat{p}^c_j)\big) \\
    &+ \big(1 - p^c_j\big) \log\big(1 - \sigma(\hat{p}^c_j)\big) \bigg]~.
    \label{eq:edge_loss}
\end{split}
\end{align}
The node loss $\mathcal{L}_{\text{nodes}}$ is the cross-entropy between the predicted node labels $\hat{v}^c$ and the ground truth labels $v^c$ for each of the $|\mathcal{V}|$ nodes across $\mathcal{C}$ classes. The edge loss $\mathcal{L}_{\text{edges}}$ is the binary cross-entropy between the predicted edge predicate logits $\hat{p}^c$ and the ground truth labels $p^c$ for each of the $|\mathcal{E}|$ edges across $\mathcal{P}$ predicate categories, where $\sigma$ denotes the sigmoid function. 
Importantly, the losses $\mathcal{L}_{\text{nodes}}$ and $\mathcal{L}_{\text{edges}}$ are computed before the information bottleneck (before nodes and edges features max-pooling). At this stage, there is an exact correspondence between the nodes and edges features and the categorical ground truth. 

Lastly, the graph latent $\mathbf{z}$ dimension is set to $\dimlatent=512$.

\smallskip
\noindent\textbf{Graph Decoder} At the decoder $\decoder$ we use $L=6$ stacked blocks, each of which performs cross-attention, self-attention, and feed-forward operations in series. The number of heads at the attention layers is set to $8$ while $d_{\text{head}}=64$.

\smallskip
\noindent\textbf{LDM Backbone} At the core of our LDM model, we use a transformer-based DiT~\cite{peebles2023dit} backbone. Specifically, we employ the DiT-S architecture, which consists of 12 stacked transformer blocks with 6 attention heads at each attention layer. 

\smallskip
\noindent\textbf{Loss Weights} 
In this section, we provide more details on the loss terms' weighting factors. 
The losses $\mathcal{L}_\text{nodes}$ and $\mathcal{L}_\text{edges}$ at the encoder are each assigned a weight of 1 in the overall optimization objective. 
For details on latent space regularization and the specific choice of hyperparameters, we point the reader to Sec.~\ref{subsec:latent_reg}.

Following \cite{im2024egtr}, the decoder loss weights are set to $\lambda_\text{obj} = 2.0$, $\lambda_\text{rel} = 15.0$, and $\lambda_\text{con} = 30.0$. The Hungarian matcher for loss computation is also adopted from \cite{im2024egtr}, using the same hyperparameters. Lastly, the RAE's regularization term hyperparameters $\beta$ and $\lambda$ are set to 0.1 and 0.0001, respectively. 

\smallskip
\noindent\textbf{SGA Baselines} We compare \ours with two baselines derived from the VidSGG literature: STTran~\cite{cong2021spatial} and DSGDetr~\cite{feng2023exploiting}, as in \cite{peddi2024towards}, and with SceneSayer~\cite{peddi2024towards}, a method specifically designed for SGA.

STTran and DSGDetr-based methods were adapted for the SGA task in \cite{peddi2024towards} to predict future relationships between fixed objects. The baseline+ variants use an anticipatory transformer to generate relationship representations for future frames auto-regressively, while the baseline++ variants further incorporate a temporal encoder to improve the model’s understanding of relationships temporal dynamics. During training, baseline+ variants focus solely on decoding anticipated relationships, while baseline++ variants employ a dual-decoding strategy to simultaneously decode both observed and anticipated relationships.

SceneSayer~\cite{peddi2024towards} is our best competitor, specifically designed for anticipating pairwise object relationships in human activities videos. It uses object-centric representations to model the temporal evolution of human-environment interactions, with two variants: SceneSayerODE, which leverages Ordinary Differential Equations (ODE), and SceneSayerSDE, which uses Stochastic Differential Equations (SDE). 
Both variants forecast the evolution of relationships between fixed objects (those observed in the last frame), hence they cannot model appearing and disappearing objects in human activities.

\begin{figure*}[t]
    \centering
    \includegraphics[width=0.98\linewidth]{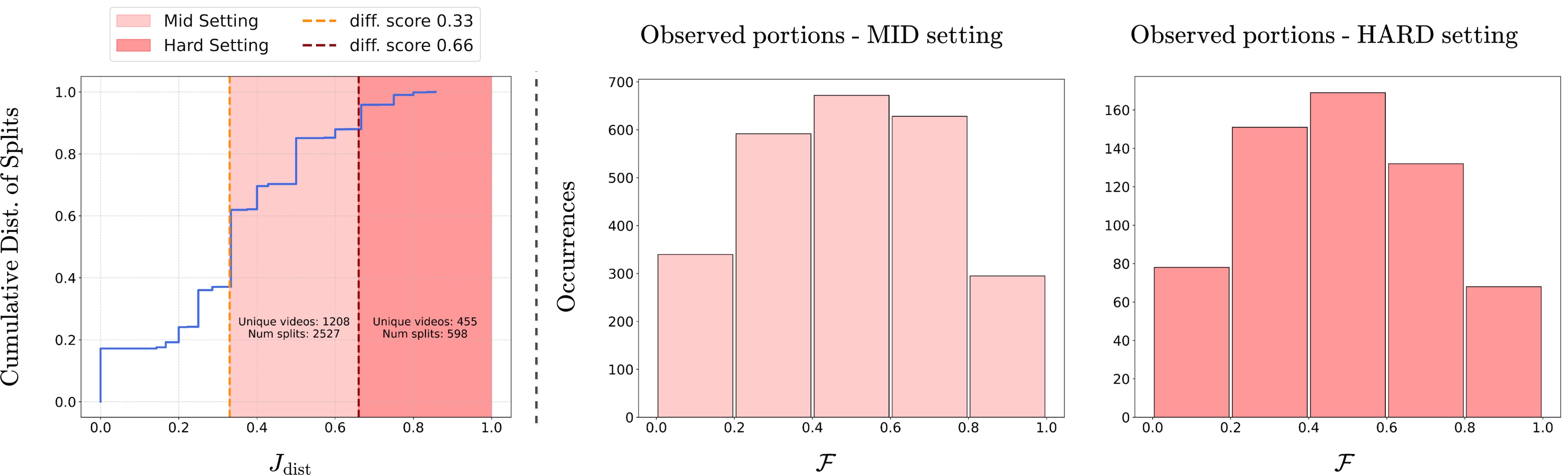}
    \caption{
    (\textit{Left}) Empirical cumulative distribution of the difficulty scores for the top-3 highest-scoring (most difficult in terms of $\jdist$) splits per test video. 
    To construct the SGA under Object Distribution Shift scenarios, we select splits with difficulty scores in the ranges  $[0.33, 0.66)$  (MID) and  $[0.66, 1]$  (HARD). 
    (\textit{Middle} and \textit{Right}) Distribution of observed video portions corresponding to anticipation splits in the MID and HARD settings. 
    \label{fig:split_diff_cdf}}
    \vspace{-4mm}
\end{figure*}

\section{SGA under Object Distribution Shift} 
Here, we provide further details on the two scenarios (MID and HARD) derived from the Action Genome test set, which consists of 1,750 videos. They define a new experimental testbed to evaluate SGA methods under real-world conditions, where object categories distribution can shift between the last observed and future frames. 

To build these scenarios, we first exclude test videos with fewer than 10 frames. For the remaining videos, we compute a \textit{difficulty score} as the Jaccard distance between the object sets in consecutive frames $F_s$ and $F_s+1$:
\begin{equation}
J_{\text{dist}} = 1 - \frac{|O_{F_s} \cap O_{F_s+1}|}{|O_{F_s} \cup O_{F_s+1}|},
\end{equation}
where $O_{F_s}$ and $O_{F_s+1}$ denote the objects in frames $F_s$ and $F_s+1$, respectively. This score quantifies the challenge of using $F_s$ as the last observed frame, with higher values indicating greater object distribution shifts.  

To ensure sufficient temporal context, we exclude the first and last three frames of each video. 
From the remaining frames, we select the top three with the highest difficulty scores to define the last observed frame in our SGA under Object Distribution Shift benchmark. This selection is based solely on the difficulty score, irrespective of where these frames appear in terms of observed video portion. 

Fig.~\ref{fig:split_diff_cdf} (\textit{left}) shows the empirical cumulative distribution of difficulty scores for these selected splits. We define difficulty-based scenarios as follows:
\begin{itemize}
    \item \textbf{MID:} video splits with scores in $[0.33, 0.66)$, comprising 2,527 anticipation splits across 1,208 unique videos.
    \item \textbf{HARD:} video splits with scores in $[0.66, 1]$, comprising 598 anticipation splits across 455 unique videos.
\end{itemize}
\smallskip
Fig.~\ref{fig:split_diff_cdf} (\textit{middle, right}) show the distribution of observed video portions corresponding to anticipation splits respectively in the MID and HARD settings. Unlike standard SGA, where the observed portion follows a fixed ratio $\mathcal{F}$, difficulty-based splits result in a nearly uniform distribution across the video rollout, indicating that object distribution shifts (both appearance of new objects and disappearance) can occur at any stage of an activity.

\begin{figure*}[t]
    \centering
    \includegraphics[width=1.\linewidth]{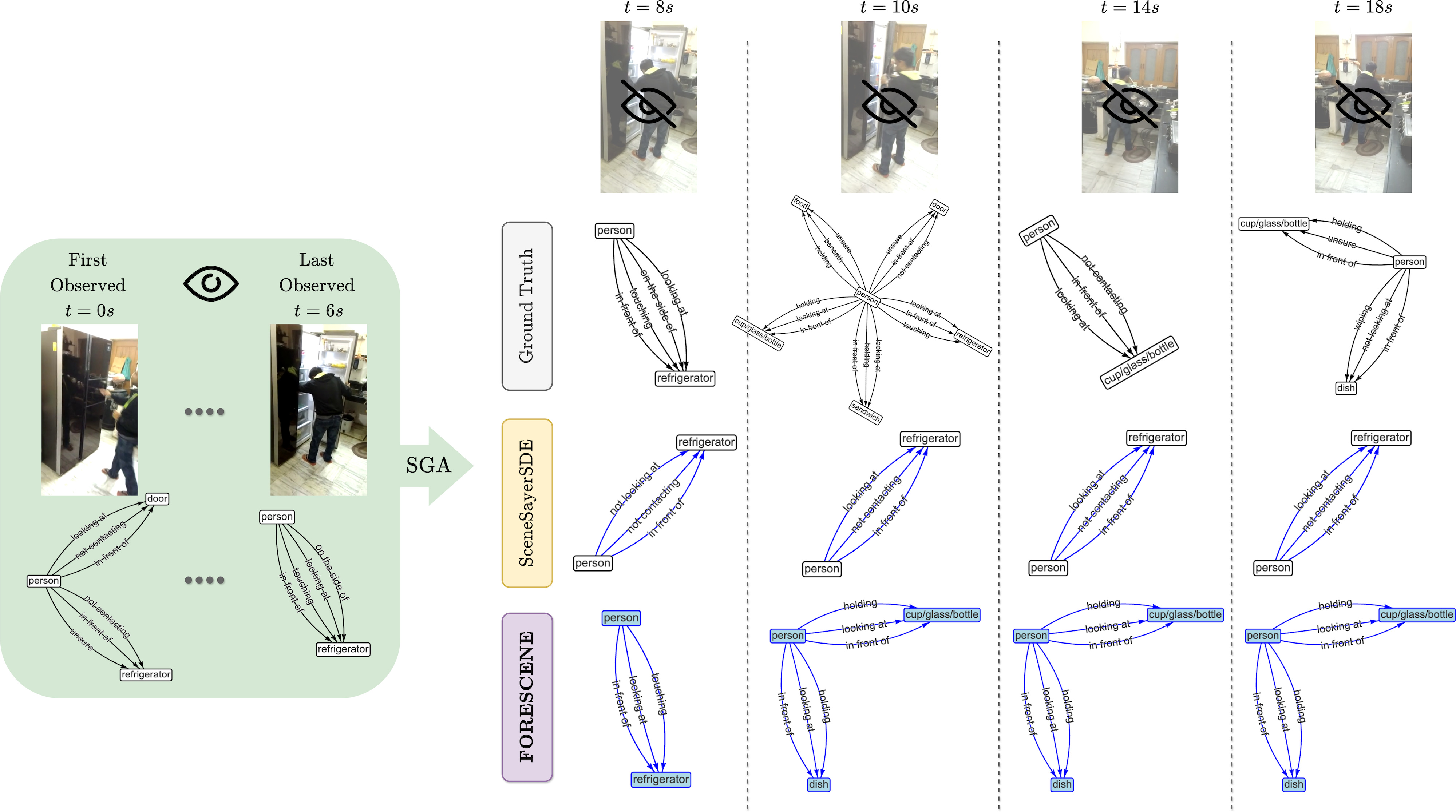}
    \caption{
    Qualitative comparison of Scene Graph Anticipation between \ours and the top-performing competitor, SceneSayerSDE~\cite{peddi2024towards}, with the observed video portion ($\mathcal{F}$) set to 0.3. 
    The examples highlight a common scenario where the object continuity assumption of previous SGA methods breaks down, hindering their applicability in real-world scenarios. 
    In contrast, \ours accurately forecasts both the appearance and disappearance of objects and their relationships over time.\label{fig:supp_qualitative}}
    \vspace{-4mm}
\end{figure*}

\section{Additional Results}
\subsection{In-Context Learning Baseline}
Since no prior work addresses SGA in an unconstrained setting where both objects and relationships evolve over time, we additionally compare \ours with a state-of-the-art Vision-Language Model (VLM) as a strong baseline for joint prediction of objects and relationships evolution. 

Specifically, we use Gemini 1.5 Pro~\cite{gemini1_5} with in-context learning, prompting it to anticipate future scene graphs based on (scene graph, visual frame) pairs from the observed video portion. 

The VLM is prompted as follows:
\texttt{You are an AI system for Scene Graph Anticipation. Given pairs of frames and their scene graphs from a video portion, predict scene graphs for subsequent unseen frames. Each scene graph consists of triplets [subject, predicate, object]. Predict scene graphs for the specified number of future frames in JSON format.}

To model the \textit{observed video portion}, the VLM is given full visual frames and attributed scene graphs (\ie, including both object and relationship annotations), providing richer visual and semantic context than standard SGA methods (including \ours), which rely solely on object bounding box features and do not exploit relationship information (see Sec. 3.2 of the main paper). This makes the VLM a strong baseline. 

To familiarize the VLM with the task and output format, we supply object and predicate category lists along with 10 in-context examples (prompt + ground truth) from the Action Genome training set. 
The model outputs scene graphs for future frames in JSON format, where each frame identifier (\eg, \texttt{frame\_20}) maps to a list of triplets \texttt{[subject, predicate, object]}, with indices referencing the specified categories. 
For each test video, the number of future frames and the corresponding JSON structure to fill are provided as input to the VLM to ensure the model generates responses in the expected format.

We compare \ours with the VLM strategy for SGA in the GAGS setting (Tab. 1 in the main paper), using the first 30\% of each video as the observed fraction ($\mathcal{F}=0.3$) to initiate the anticipation process. 
For Object Discovery, the VLM achieves an average Jaccard similarity ($\jsim$) of 0.57, while \ours outperforms it with scores of 0.61 (r=1) and 0.68 (r=10). 
For Triplets Evaluation, the VLM strategy lags behind ours R@10 With Constraint (VLM: 30.3 vs \ours: 36.4).

\subsection{Qualitative Results}
In Fig.~\ref{fig:supp_qualitative}, we provide a qualitative comparison of our approach, \ours, with the leading competitor, SceneSayerSDE\cite{peddi2024towards}, for SGA. 
Specifically, we set the observed video portion to $\mathcal{F}=0.3$ and compare the predicted sequence of graphs against the ground truth sequence.

The example illustrates how previous SGA methods struggle when the assumption of object continuity is not satisfied, which we argue is a common condition in many (dynamic) real-world scenarios. For instance, SceneSayerSDE persistently maintains the nodes \textit{person} and \textit{refrigerator} throughout the entire forecasted graph sequence, disregarding the interaction with other objects not active in the last observed frame. 

In contrast, \ours dynamically and accurately predicts the emergence and disappearance of objects and their relationships over time. 
In this example, \ours figures out that after the \textit{person} interacts with the \textit{refrigerator}, they are likely to retrieve items such as a \textit{dish} for eating and potentially a \textit{cup}, \textit{glass}, or \textit{bottle} for drinking.

\end{document}